\pdfoutput=1

\documentclass[11pt]{article}

\usepackage[final]{acl}

\usepackage{amsmath,amsfonts,bm}

\def\eqref#1{equation~\ref{#1}}

\def\1{\bm{1}}

\DeclareMathAlphabet{\mathsfit}{\encodingdefault}{\sfdefault}{m}{sl}
\SetMathAlphabet{\mathsfit}{bold}{\encodingdefault}{\sfdefault}{bx}{n}

\def\gD{{\mathcal{D}}}
\def\gE{{\mathcal{E}}}

\def\gS{{\mathcal{S}}}

\def\gV{{\mathcal{V}}}

\DeclareMathOperator*{\argmax}{arg\,max}

\usepackage{times}
\usepackage{latexsym}
\usepackage{mdframed}
\usepackage[T1]{fontenc}

\usepackage[utf8]{inputenc}

\usepackage{microtype}

\usepackage{inconsolata}

\usepackage{caption}
\usepackage{subcaption}
\usepackage{enumitem}
\usepackage{flushend}
\usepackage{balance}
\usepackage{lineno}
\usepackage{url}
\usepackage{amsmath,amssymb,amsfonts}
\usepackage{algorithm}
\usepackage{algorithmic}
\usepackage{graphicx}
\usepackage{bbding}
\usepackage{textcomp}
\usepackage{xcolor}
\usepackage{hyperref}
\usepackage{colortbl}
\usepackage{amsthm}
\usepackage{float}
\usepackage{multirow}
\usepackage{multicol}
\usepackage{color}
\usepackage{bm}
\usepackage{bbm}

\usepackage{pifont}

\newcommand{\bX}{\mathbf{X}}

\newcommand{\bA}{\mathbf{A}}

\newcommand{\bH}{\mathbf{H}}

\usepackage{array}
\newcolumntype{L}[1]{>{\raggedright\let\newline\\\arraybackslash\hspace{0pt}}m{#1}}
\newcolumntype{C}[1]{>{\centering\let\newline  \\\arraybackslash\hspace{0pt}}m{#1}}
\newcolumntype{R}[1]{>{\raggedleft\let\newline \\\arraybackslash\hspace{0pt}}m{#1}}

\newcommand{\mypara}[1]{{\smallskip \noindent \bf #1}\hspace{0.1in}}

\usepackage[utf8]{inputenc} 
\usepackage[T1]{fontenc}    
\usepackage{hyperref}       
\usepackage{url}            
\usepackage{booktabs}       
\usepackage{amsfonts}       
\usepackage{nicefrac}       
\usepackage{microtype}      
\usepackage{xcolor}         
\usepackage{comment}
\usepackage{arydshln}

\title{From Cross-Task Examples to In-Task Prompts: A Graph-Based Pseudo-Labeling Framework for In-context Learning}

\author{Zihan Chen, Song Wang, Xingbo Fu, Chengshuai Shi, Zhenyu Lei, Cong Shen, Jundong Li\\
Department of ECE, University of Virginia, Charlottesville, VA, USA\\
\texttt{\{brf3rx,sw3wv,xf3av,cs7ync,vjd5zr,cs7dt,jl6qk\}@virginia.edu}}

\begin{document}
\maketitle
\begin{abstract}
The capability of in-context learning (ICL) enables large language models (LLMs) to perform novel tasks without parameter updates by conditioning on a few input-output examples. However, collecting high-quality examples for new or challenging tasks can be costly and labor-intensive. In this work, we propose a cost-efficient two-stage pipeline that reduces reliance on LLMs for data labeling. Our approach first leverages readily available cross-task examples to prompt an LLM and pseudo-label a small set of target task instances. We then introduce a graph-based label propagation method that spreads label information to the remaining target examples without additional LLM queries. The resulting fully pseudo-labeled dataset is used to construct in-task demonstrations for ICL. This pipeline combines the flexibility of cross-task supervision with the scalability of LLM-free propagation. Experiments across five tasks demonstrate that our method achieves strong performance while lowering labeling costs. Our code is available at~\href{https://github.com/Chen-1031/Cross-Task-ICL}{https://github.com/Chen-1031/Cross-Task-ICL}.

\end{abstract}

\section{Introduction}\label{sec:intro}

Large language models (LLMs) have demonstrated impressive capabilities across a wide range of natural language processing tasks~\cite{zhao2023survey,chang2024survey}, including semantic parsing~\cite{li2021mtop,wolfson2020break} and commonsense reasoning~\cite{talmor-etal-2019-commonsenseqa,zellers2019hellaswag,lei2025learning,lei2025harnessing}. However, the substantial computational cost of retraining or fine-tuning these models limits their practicality for novel tasks~\cite{hu2021lora,liu2022few,zaken2022bitfit}. Fortunately, LLMs possess an emergent ability known as In-Context Learning (ICL)~\cite{wang2024large,wang2024mixture,chenmaple}, wherein the model can perform new tasks by conditioning on a few input-output pairs (i.e., demonstrations) during inference, without updating model parameters~\cite{brown2020language}.

\begin{figure}[t]
\centering

\begin{minipage}{\linewidth}
  \centering
  \includegraphics[width=\linewidth]{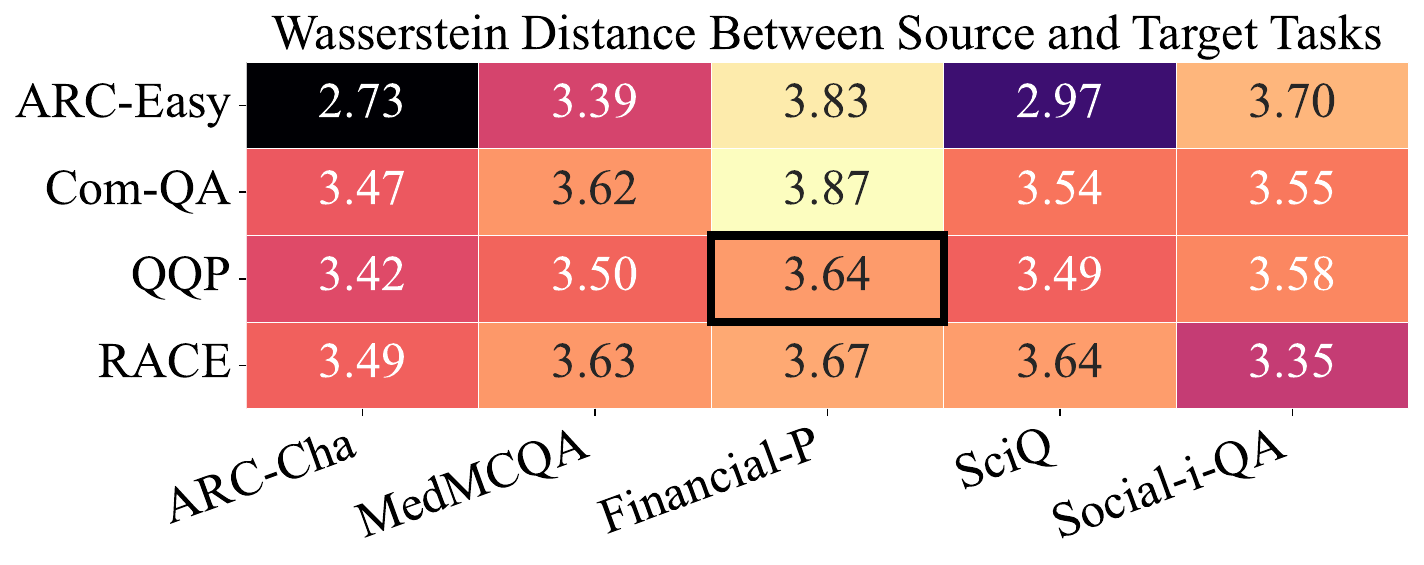}
\end{minipage}


\begin{minipage}{\linewidth}
  \centering
  \small
    \tabcolsep=6pt
  \begin{tabular}{ll}
    \toprule
    \textbf{Task} & \textbf{Label Space} \\
    \midrule
    QQP & \{duplicate, not duplicate\} \\
    Financial-Phrasebank & \{positive, neutral, negative\} \\
    \bottomrule
  \end{tabular}
\end{minipage}

\caption{(Top) Wasserstein distance between source (column) and target (row) task example embeddings. (Bottom) Examples of task label spaces.}
\label{fig:teaser}
\vspace{-0.2in}
\end{figure}

Despite its promise, the effectiveness of ICL heavily relies on high-quality labeled examples for the target task. For novel or data-scarce tasks, recent work has explored using LLMs with zero-shot prompts~\cite{zhang2025large,wan2024tnt,shi2024efficient} or relying on human annotators~\cite{mikulova2023quality,wang2024human} to obtain pseudo-labeled examples, which are then used as demonstrations for ICL. Yet both approaches have drawbacks: LLMs can be unreliable on unfamiliar tasks, while human annotation introduces additional time and labor costs~\cite{su2022selective}.
To overcome this, recent efforts have turned to leveraging well-established, high-resource source tasks to construct demonstrations~\cite{tanwar2023multilingual,raffel2020exploring}. When the examples used for ICL are drawn from a different task than the target, the setting is referred to as cross-task ICL. For example, \citet{chatterjee2024language} select examples from source tasks based on embedding similarity and demonstrate that such cross-task examples can significantly improve ICL performance, highlighting cross-task ICL as a promising approach for pseudo-labeling.

However, as illustrated in Figure~\ref{fig:teaser}, the data distributions of source and target tasks often differ significantly, and the label spaces can be misaligned, even between tasks with similar distributions. This raises a key limitation: selecting cross-task examples based solely on embedding similarity is insufficient for reliable pseudo-labeling of target samples.

To address this, we draw inspiration from the graph mining literature, which shows that structural properties of graphs can generalize across domains even when feature spaces are heterogeneous ~\cite{qiu2020gcc,leskovec2005graphs,hamilton2017graphsage}. Based on this, we propose GraphSim, a graph-based example selection method that augments text embeddings with structural information captured through graph aggregation—a process where each node's representation is updated by aggregating information from its neighbors. These structure-aware embeddings yield more robust similarity metrics across tasks, allowing for better example selection.
To further address the label space mismatch across tasks, we propose GLIP (Graph-based Label Information Propagation), a label information propagation framework that uses a small set of pseudo-labeled target examples (obtained via GraphSim and an LLM) to infer labels for the remaining unlabeled target examples. The resulting pseudo-labeled target set can then be used as high-quality demonstrations for in-context learning.
Notably, our pipeline is cost-efficient and adaptable: it only requires a small number of LLM calls for pseudo-labeling and leverages lightweight graph-based propagation for the rest, making it practical for real-world deployment on novel tasks. Our contributions can be summarized as follows:
\begin{itemize}[left=0pt]
    \item \textbf{Problem Formulation}: We propose a novel problem formulation that utilizes examples from high-resource source tasks to pseudo-label examples from a novel target task. This enables in-context learning without requiring extensive manual annotation or large-scale LLM inference on the target task.
    \item \textbf{Methodological Innovation}: We introduce a two-stage graph-based pipeline that addresses both cross-task example selection and label propagation. First, we propose GraphSim, a structure-aware similarity metric for selecting source examples that are more transferable across tasks. Second, we design GLIP, which efficiently propagates labels from a few LLM-labeled target examples to the rest of the target dataset, mitigating label space misalignment.
    \item \textbf{Empirical Validation}: Comprehensive experiments on five target tasks with five different-sized LLMs demonstrate that our method outperforms existing cross-task baselines and approaches the performance of in-task upper bounds with light reliance on LLMs, highlighting both its effectiveness and efficiency.
\end{itemize}

\section{Related Works}

\mypara{In-Context Learning.}  
In-context learning (ICL) \cite{brown2020language} equips large language models (LLMs) with the ability to leverage a handful of input-output demonstrations for reasoning. ICL has proven remarkably successful in handling complex tasks, including summarization~\cite{jain2023multi,baek2024revisiting} and question answering~\cite{lee2024can}.
To improve ICL performance, many works have explored adaptive strategies for selecting effective demonstrations~\cite{lu2021fantastically,zhao2021calibrate,ye2023compositional,chen2024fastgas}, such as retrieving semantically similar examples~\cite{liu2021makes}. However, these methods typically assume that the demonstrations and the input come from the same task (i.e., in-task ICL) and rely on access to a large pool of labeled examples. This assumption limits their applicability in low-resource or novel task settings, where collecting high-quality annotations is costly and impractical.

\mypara{Cross-task In-Context Learning.}
Due to the large disparity in annotation availability between novel tasks and well-established ones, leveraging high-resource source tasks to improve performance on low-resource target tasks has become an appealing direction. Prior work has explored cross-task ICL in various settings, such as cross-lingual tasks~\cite{tanwar2023multilingual}, multi-task learning~\cite{zhang2022task,raffel2020exploring}, and prompt generation for downstream tasks~\cite{zou2023meta}. More recently, \citet{chatterjee2024language} demonstrated that LLMs can benefit significantly from cross-task ICL prompts and showed the potential of generating pseudo-labels for in-task examples. 
However, their method relies on semantic similarity for example selection and labels only a small subset of target examples, leaving most unlabeled data unused. In contrast, our cost-efficient pipeline combines LLM-based pseudo-labeling of a small seed set with graph-based label information propagation, enabling scalable construction of high-quality in-task demonstrations.

.

\begin{figure*}[!t] 
\label{fig:framework}
\begin{center}
\centerline{\includegraphics[width=\textwidth]{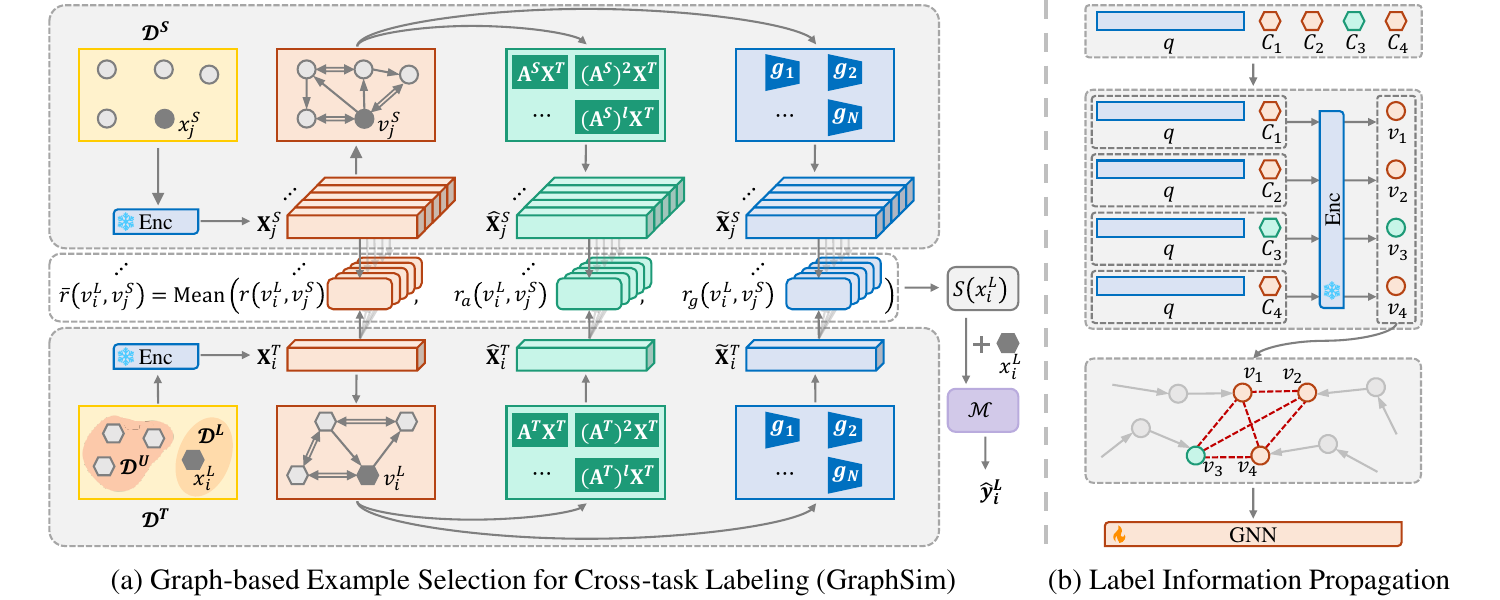}}
\caption{Overview of our proposed pipeline for cross-task pseudo-labeling. We first use (a) \textbf{GraphSim} to select relevant examples from the source task to pseudo-label a small set of target task examples $\gD^L$ via ICL. Then, we apply (b) \textbf{GLIP}, a graph-based label propagation method, to infer labels for the remaining unlabeled target samples $\gD^U$. The resulting fully pseudo-labeled target set is used to construct in-task examples for in-task ICL}
\label{figure:pretrain}
\end{center}
\vskip -0.2in
\end{figure*}

\section{Preliminaries}

\subsection{Graph Neural Networks}
Let $\mathcal{G} = (\mathcal{V}, \bA, \bX)$ denote an attributed graph with a set of nodes $\mathcal{V} = \{ v_1, v_2, \cdots, v_{|\mathcal{V}|} \}$. 
$\bX \in \mathbb{R}^{|\mathcal{V}| \times d}$ is the feature matrix where each row $\bX_i \in \mathbb{R}^d$ is the $d$-dimensional feature vector of node $v_i \in \mathcal{V}$.
$\bA \in \{0, 1\}^{|\mathcal{V}| \times |\mathcal{V}|}$ is the adjacency matrix, where each entry $\bA_{ij}=1$ if nodes $v_i$ and $v_j$ are connected by an edge; otherwise, $\bA_{ij}=0$.
GNNs typically follow a message-passing framework~\cite{kipf2016gcn, velivckovic2017gat}, where each node iteratively aggregates information from its neighbors. At the $l$-th layer, node $v_i$'s representation $\bH^{(l)}_i$ is updated as:
\begin{equation} \label{gnn}
    \bH^{(l)}_i = g^{(l)}(\bH^{(l-1)}_i, \{\bH^{(l-1)}_j: v_j \in \mathcal{N}(v_i)\}; \theta^{(l)}),
\end{equation}
where $\mathcal{N}(v_i)$ denotes the neighbors of $v_i$, and $g^{(l)}$ is the aggregation function with parameters $\theta^{(l)}$. 
We initialize $\bH^{(0)}_i$ as the node feature, i.e., $\bH^{(0)}_i=\bX_i$.

\subsection{In-Contex Learning}
In-Context Learning (ICL) enables LLMs to perform a new task simply by conditioning on a few input-output examples without any fine-tuning~\cite{brown2020language}. Given a prompt composed of $k$ examples $\{(x_j, y_j)\}_{j=1}^k$ and a new query input $x_{\text{query}}$, the model generates a prediction $\hat{y}_{\text{query}}$ by autoregressively decoding the next token(s):
\[\hat{y}_{\text{query}}=\mathcal{M}((x_i,y_1),...,(x_k,y_k),x_{\text{query}}),\]
where $\mathcal{M}$ is the frozen LLM. The effectiveness of ICL largely depends on the quality and relevance of the selected examples~\cite{liu2021makes}.

\section{Methodology}

\subsection{Problem Setup}
We formulate each input-output pair $(x, y)$ as a multiple-choice question, where $x = (q, \{C_1, \ldots, C_n\})$ consists of a query and $n$ candidate choices, and $y$ is the correct answer.

Let $\mathcal{D}^S = \{(x_i^S, y_i^S)\}_{i=1}^{|\mathcal{D}^S|}$ be a labeled source dataset, and let $\mathcal{D}^T = \{(x_i^T, y_i^T)\}_{i=1}^{|\mathcal{D}^T|} = \mathcal{D}^L \cup \mathcal{D}^U$ be the target dataset, with $\mathcal{D}^L$ denoting a small subset used for cross-task in-context labeling, and $\mathcal{D}^U$ the remaining unlabeled examples. 
For each $x^L \in \mathcal{D}^L$, we use a pretrained language model $\mathcal{M}$ to generate a pseudo-label $\hat{y}^L$ conditioned on selected source examples $\mathcal{S}(x^L) \subseteq \mathcal{D}^S$:
\begin{equation}\label{eqn:icl}
\hat{y}^L = \mathcal{M}(\mathcal{S}(x^L), x^L).
\end{equation}
To avoid labeling the full target set with LLMs, we propagate label information from $\mathcal{D}^L$ to $\mathcal{D}^U$ using an LLM-free algorithm. The resulting pseudo-labeled target set is then used to construct in-task examples for ICL.

\subsection{Graph-based Example Selection for Cross-task Labeling (GraphSim)}\label{sec:selection}
In this section, we propose a graph-based method, GraphSim, to select cross-task examples for pseudo-labeling. 
Traditional ICL selects examples based on embedding similarity, assuming query and examples come from the same task. This assumption fails in cross-task settings, where data distributions and label spaces often differ (Figure~\ref{fig:teaser}).

To address this, we leverage structural patterns in data via graph-based modeling. Our approach GraphSim is motivated by findings in graph mining, which suggest that structural properties of graphs can generalize across domains, even when feature spaces are heterogeneous~\cite{qiu2020gcc,leskovec2005graphs}. Specifically, we independently construct task-specific graphs for both source and target tasks. We then apply graph-based aggregation to enrich the embeddings with structural information. These structure-aware embeddings better capture task-level semantics, enabling more meaningful similarity computations across tasks. To further enhance representativeness, GraphSim incorporates a set of randomly initialized GNNs to introduce diverse views during aggregation, enhancing the cross-task example selection.

Specifically, given the source task dataset $\mathcal{D}^S$, we construct a graph \(\mathcal{G}^S = (\mathcal{V}^S, \bA^S, \bX^S)\) to model the relationships among samples, where each node $v^S_i\in \mathcal{V}^S$ corresponds to a sample $x^S_i \in \mathcal{D}^S$ whose feature vector is obtained through a pre-trained text encoder, i.e., $\bX^S_i = \mathtt{Enc}(x^S_i)$. 
To obtain $\bA^S$, we connect each node to its top-$k$ most relevant neighbors based on a pairwise relevance score. Specifically, the relevance between two samples is computed as the cosine similarity of their encoded representations:
\begin{equation}
r(v^S_i, v^S_j) =   \mathrm{cos}( \bX^S_i, \bX^S_j).
\end{equation}  
Afterwards, each entry $\bA^S_{ij} \in \bA^S$ of the source graph is computed as:
\begin{equation}
\bA^S_{ij} = \left\{
\begin{array}{c l}
1, & \text{if} \; r(v^S_i, v^S_j) \in \text{Top-}k \left\{r(v^S_i, v^S_j)\right\},\\
0, & \text{otherwise},
\end{array} 
\color{white}\right\}
\end{equation}
ensuring that only the $k$ most relevant connections are retained for each node. 
Similarly, we can construct the graph $\mathcal{G}^T = (\mathcal{V}^T, \bA^T, \bX^T)$ for the target task using the target dataset $\gD^T$.

To incorporate structural information into the node representations, we apply two types of aggregation: (1) Adjacency-based aggregation, and (2) GNN-based aggregation.

The adjacency-based aggregation captures multi-hop neighborhood information by applying powers of ${\bA^S}$ to $\bX^S$ and concatenating the results:
\[\hat{\bX}^S=[{\bA^S}\bX^S || \left( {\bA^S}\right)^2 \cdot \bX^S||\cdots||\left( {\bA^S}\right)^l\bX^S],\]
where $l$ controls the number of hops (i.e., the neighborhood depth). This allows each node to gather information from its $l$-hop neighbors and encode structural patterns beyond immediate connections.

Beyond adjacency-based aggregation, we further capture structural information using a set of randomly initialized GNNs. The key intuition is that if two nodes exhibit similar structural patterns, their aggregated representations from a GNN will also be similar. 
To achieve this without requiring training, we initialize a collection of GNNs $\{g_1, g_2, \ldots, g_N\}$, each with a different number of layers and independently randomized parameters. This design offers two main advantages:
(i) GNNs with varying depths and parameterizations capture structural features from diverse perspectives;
(ii) Random initialization avoids training overhead, making our method efficient, easily scalable, and broadly applicable across tasks.
The GNN-based aggregation is then formulated as:
\[\tilde{\bX}^S=[g_1(\bA^S, \bX^S)||\cdots||g_N(\bA^S, \bX^S)].\]

With the proposed two aggregation methods, we generate augmented representations for the source and target tasks separately as follows:
\[\mathcal{X}^S=\{(\bX^S,\hat{\bX}^S,\tilde{\bX}^S)\},\]
\[\mathcal{X}^T=\{(\bX^T,\hat{\bX}^T,\tilde{\bX}^T)\},\]
where each tuple combines the original, adjacency-based, and GNN-based representations of a sample.

We then compute the cross-task similarity between nodes $v^T_i \in\gV^T$ and $v^S_j \in\gV^S$ by averaging the similarities of each type of embedding:
\begin{align*}
    \bar{r}(v^T_i, v^S_j)=&\mathtt{Mean}(r(v^T_i, v^S_j),\\ &r_a(v^T_i, v^S_j),r_g(v^T_i, v^S_j)),
\end{align*}
where $\mathtt{Mean}(\cdot, \cdot, \cdot)$ denotes the average operation.
$r_a$ and $r_g$ are similarity scores based on adjacency- and GNN-augmented embeddings, respectively. 
Averaging the separately computed similarities offers computational efficiency and helps mitigate the curse of dimensionality that may arise from direct concatenation.

Using the cross-task similarity $\bar{r}$, we select the source examples $\gS(x^L_i)\subseteq\gD^S$ for $x^L_i\in \gD^L$ as:
\[\gS(x^L_i) = \{x^S_j|\bar{r}(x^S_j, x^L_i) \in \text{Top-}K (\{\bar{r}(x^S, x^L_i)\})\}.\]
Following Equation~\ref{eqn:icl}, we use the LLM's prediction $\hat{y}_i^L$ to construct a pseudo-labeled set $\bar{\mathcal{D}}^L =\{(x_i^L,\hat{y}_i^L)\} $, which provides label information for subsequent in-task information propagation.

\subsection{Label Information Propagation (GLIP)}\label{sec:glip}




Prior work shows that increasing in-context examples generally boosts ICL performance by providing richer task-specific signals~\cite{agarwal2024many}. While labeling the entire target set $\gD^T$ via cross-task ICL is possible, it incurs high cost as the number of LLM calls grows linearly with data size. To reduce this cost, we adopt graph-based semi-supervised learning, where labels are propagated from a small labeled set. However, applying this to QA tasks is nontrivial—labels in multiple-choice QA must be invariant to choice order.


To address these challenges while leveraging the strengths of graph-based learning, we propose GLIP, a tailored graph construction method with two types of edges and a GNN trained to perform label information propagation.

We begin with graph construction. To ensure invariance to choice order, we encode QA examples in a query-choice pairwise way. Specifically, for each pseudo-labeled example $(x,\hat{y})=(q, \{C_1,C_2,...,C_n\}, \hat{y}) \in\bar{\mathcal{D}}^L$, we generate $n$ nodes with features $\{\mathtt{Enc}([q,C_1]),...,\mathtt{Enc}([q,C_n])\}$. The multiple-choice QA task is thus transformed into a multi-node binary classification problem, where each node receives a label of 1 (correct) or 0 (incorrect). For example, if $\hat{y}=C_3$, then we label the node with feature $\mathtt{Enc}([q,C_3])$ as 1, and the remaining nodes are labeled as 0. For unlabeled examples in $\gD^U$, we construct nodes in the same way but leave their labels unassigned.

We then define two types of edges to capture task-specific structure: (i) Similarity-based positive edges $\gE_{pos}$: Constructed using the relevance scores between query-choice pairs, as described in Section~\ref{sec:selection}. These edges \textit{only} connect nodes from different queries to model semantic relationships across both labeled and unlabeled samples, which is critical for pseudo-labeling unlabeled samples. (ii) Mutual exclusion negative edges $\gE_{neg}$: These encode Mutual Exclusion Constraints (MEC)~\cite{su2022zlpr} by connecting nodes within the same query. Since only one choice can be correct, these edges penalize configurations where multiple nodes for the same question are labeled as correct (i.e., 1) during training, enforcing consistency in the multi-choice setting.

Following the construction of nodes and edges, we obtain a graph $\mathcal{G}$ composed of labeled and unlabeled nodes. To perform semi-supervised learning, we train a graph neural network (GNN) $\tilde{g}$ on $\mathcal{G}$ by minimizing the objective $\mathcal{L}=\mathcal{L}_{CE}+\lambda \mathcal{L}_{MEC}$, where $\mathcal{L}_{CE}$ is the standard cross-entropy loss over labeled nodes, and $\mathcal{L}_{MEC}$ is a MEC loss designed to reduce the similarity between nodes connected byy $\gE_{neg}$:
\[\mathcal{L}_{MEC}=-\frac{1}{|\gE_{neg}|}\sum_{i,j\in\gE_{neg}} \langle \mathbf{h}_i, \mathbf{h}_j \rangle,\]
where $\mathbf{h}_i$ and $\mathbf{h}_j$ are the output embeddings of nodes $i$ and $j$ from the GNN $\tilde{g}$, respectively.

Once trained, $\tilde{g}$ is used to predict labels for unlabeled nodes in $\mathcal{G}$. For each unlabeled input $x\in \gD^U$, we determine its predicted answer by selecting the choice with the highest logit:
\[\hat{y}=\argmax_{i\in[n]}\tilde{g}(Enc([q,C_i]).\]

After combining the pseudo-labeled sets from the cross-task labeling stage and the graph-based label propagation, we obtain a fully pseudo-labeled target dataset $\bar{\mathcal{D}}^T = \bar{\mathcal{D}}^L \cup \bar{\mathcal{D}}^U$.  This augmented set can then be used in a traditional similarity-based in-context learning pipeline for final inference on the target task.

\section{Experiment}
\subsection{Experiment setup}
\textbf{Datasets and experimental setup.}
We follow the same source and target dataset setting as in \citet{chatterjee2024language}. Detailed task descriptions for both source and target tasks are provided in Appendix~\ref{sec:appendix-dataset}. We focus on the single-source-task scenario and select the best-performing source task for each target task based on the results reported in \citet{chatterjee2024language}\footnote{Identifying the optimal source remains an open challenge. One promising direction is to select source tasks based on the similarity of final-layer hidden states between source and target task definitions \cite{chatterjee2024language}.}. The source-target task pairs used in our experiment are listed in Table~\ref{tab:pair}.

For consistency, we standardize dataset sizes: we sample 10,000 examples from each source task ($|\gD^S|=10,000$) and construct a 500-example pool for each target task ($|\gD^T|=500$), among which 100 examples are used for cross-task labeling and the remaining 400 are used for label propagation. An additional disjoint set of 500 target examples is reserved for final inference evaluation.

We use Sentence-BERT~\cite{reimers2019sentence} as the text encoder $\mathtt{Enc}(\cdot)$. Unless otherwise specified, we set 
$k=20$ for graph construction and $l=2$ for adjacency-based aggregation. For GNN-based aggregation, we initialize $N=4$ randomly initialized GCNs~\cite{kipf2016gcn}: two with a single layer and two with two layers, each with hidden size 128. For label information propagation, we use a two-layer GAT~\cite{velivckovic2017gat} with hidden size 64 as the backbone model. We train it for 25 epochs using the Adam optimizer with a learning rate of 0.005. The loss balancing coefficient is set to 
$\lambda=0.4$.
For LLM-based cross-task labeling, we follow \citet{chatterjee2024language} and evaluate with LLaMA2-7B, LLaMA2-13B~\cite{touvron2023llama}, and GPT-3.5~\cite{gpt3.5turbo_doc}. Additional results using LLaMA3-8B~\cite{llama3modelcard} and GPT-4o~\cite{hurst2024gpt} are presented in the Appendix \ref{sec:morellm}.

\begin{table}[H]
\caption{Source-Target task pairs used in the experiment.}
\centering
\small
\tabcolsep = 4pt
\begin{tabular}{l c c}
\toprule[1.2pt]
    \textbf{LLM} & \textbf{Target Task} & \textbf{Source Task} \\ 
    \midrule
    \multirow{5}{*}{\shortstack{\textbf{LLaMA2-7B} \\ \textbf{LLaMA3-8B}}} & ARC-Challenge  &  ARC-Easy    \\ 
    & MedMCQA	  &Commonsense-QA     \\ 
    & Financial-Phrasebank  &SST2    \\ 
    & SciQ  &Commonsense-QA    \\ 
    & Social-i-QA  &RACE     \\ \midrule
    \multirow{5}{*}{\textbf{LLaMA2-13B}} & ARC-Challenge  &  ARC-Easy    \\ 
    & MedMCQA	  &RACE     \\ 
    & Financial-Phrasebank  &QQP    \\ 
    & SciQ  &Commonsense-QA     \\ 
    & Social-i-QA  &RACE     \\ \midrule
    \multirow{5}{*}{\shortstack{\textbf{GPT-3.5} \\ \textbf{GPT-4o}}} & ARC-Challenge  &RACE     \\ 
    & MedMCQA	  &BoolQ     \\ 
    & Financial-Phrasebank  &AG-news    \\ 
    & SciQ  &RACE     \\ 
    & Social-i-QA  &RACE    \\ 
\bottomrule[1.2pt]
\end{tabular}
\vspace{-0.1in}
\label{tab:pair}
\end{table}

\noindent \textbf{Baselines.} 
We compare our pipeline against three types of baselines: i) \textbf{Zero-shot}: The LLM is prompted with only the task instruction. ii) \textbf{Cross-task ICL}: Includes EmbSim and GraphSim, which retrieve examples from a source task. EmbSim only uses embedding similarity, while GraphSim applies our graph-based method in Section~\ref{sec:selection}. iii) \textbf{In-task ICL}: Includes $\mathcal{L}_{\text{LLM}}$, GLIP, and Oracle. These differ in how they construct the labeled target pool $\gD^T$. $\mathcal{L}_{\text{LLM}}$ uses labeled $\gD^L$ to perform in-task ICL and pseudo-label $\gD^U$; GLIP applies our proposed method in Section~\ref{sec:glip}) to label $\gD^U$; and Oracle represents an idealized case where the entire $\gD^T$ is assumed to be gold-labeled and directly used as the demonstration pool. A summary of the differences between these methods is provided in Table~\ref{tab:bsl}.

\begin{table}
\caption{Source of examples used by different baselines. \textcolor{red!70}{Red} lines indicate cross-task ICL settings, and \textcolor{cyan!70}{Blue} lines indicate in-task ICL settings.}
\centering
\small
\tabcolsep = 4pt
\begin{tabular}{l c c c }
\toprule[1.2pt]
    \textbf{Method} & \textbf{$\gD^S$} & \textbf{$\gD^L$}& \textbf{$\gD^U$} \\ 
    \midrule
    \textbf{Zero-shot} & \XSolidBrush  & \XSolidBrush   & \XSolidBrush   \\ 
    \rowcolor{red!10}\textbf{EmbSim} &Ground Truth   & \XSolidBrush & \XSolidBrush   \\ 
    \rowcolor{red!10}\textbf{GraphSim} &Ground Truth   & \XSolidBrush & \XSolidBrush \\ 
    \rowcolor{cyan!10}\textbf{$\mathcal{L}_{LLM}$} & \XSolidBrush  &Ground Truth   &In-task ICL   \\ 
    \rowcolor{cyan!10}\textbf{GLIP} & \XSolidBrush &Ground Truth   &GLIP   \\ 
    \rowcolor{cyan!10}\textbf{Ours} & \XSolidBrush &GraphSim   &GLIP   \\ 
    \rowcolor{cyan!10}\textbf{Oracle} & \XSolidBrush  &Ground Truth   &Ground Truth   \\ 
\bottomrule[1.2pt]
\end{tabular}
\vspace{-0.1in}
\label{tab:bsl}
\end{table}

\subsection{Main Result}
Table~\ref{tab:main-results} presents the performance of our proposed pipeline and six baselines across five target tasks under one-shot and four-shot in-context learning settings. We highlight the following key observations:
(i) \textbf{Graph-based similarity improves cross-task in-context learning.} Both EmbSim and GraphSim operate in the cross-task ICL setting. Across all tasks, GraphSim consistently outperforms EmbSim, demonstrating that augmenting input representations with structural information, shared across different domains, helps identify more relevant examples for cross-task labeling. This leads to higher-quality pseudo-labels for the target task.
(ii) \textbf{Label information propagation outperforms LLM-based pseudo-labeling.} $\mathcal{L}_{LLM}$ and GLIP use the same golden-labeled set $\gD^L$, but differ in how they generate labels for $\gD^U$: $\mathcal{L}_{LLM}$ uses in-task ICL with the LLM, while GLIP applies our proposed graph-based propagation method. GLIP consistently outperforms $\mathcal{L}_{LLM}$ and achieves performance closer to the Oracle, suggesting that label propagation is more reliable --- particularly for challenging tasks where LLM-generated labels can be noisy and unreliable.
(iii) \textbf{Our full pipeline is both effective and cost-efficient.} The combination of graph-based cross-task labeling and label propagation results in high-quality pseudo-labeled target sets, enabling strong ICL performance on novel tasks. Our method significantly reduces reliance on LLM inference during labeling, making it more efficient and practical in real-world, resource-constrained settings.

\begin{table*}
\caption{Performance comparison of our pipeline and six baselines on five target tasks under one-shot and four-shot in-context learning settings. Dashed lines separate cross-task ICL results from in-task ICL results. Bold font highlights the best performance within each category, while underlined values indicate the second-best results within the in-task ICL group for clearer analysis.}
\centering
\small
\tabcolsep = 4pt
\begin{tabular}{l c c c c c c c c c c}
\toprule[1.2pt]
    \multirow{2}{*}{\textbf{Method}} & \multicolumn{2}{c}{\textbf{ARC-Challenge}} & \multicolumn{2}{c}{\textbf{MedMCQA}}& \multicolumn{2}{c}{\textbf{Financial-Phrasebank}}& \multicolumn{2}{c}{\textbf{SciQ}}& \multicolumn{2}{c}{\textbf{Social-i-QA}} \\ 
    \cmidrule(lr){2-3} \cmidrule(lr){4-5} \cmidrule(lr){6-7} \cmidrule(lr){8-9} \cmidrule(lr){10-11}
     &$K=1$ &$K=4$	&$K=1$ &$K=4$ &$K=1$ &$K=4$ &$K=1$ &$K=4$ &$K=1$ &$K=4$  \\ 
    \midrule
    \rowcolor{gray!20} \multicolumn{11}{c}{\textbf{\emph{LLaMA2-7B}}}\\ 
    \textbf{Zero-shot} &4.6  &4.6 &4.2 &4.2 &34.1 &34.1 &8.0 &8.0 &41.1 &41.1\\ 
    \rowcolor{red!10}\textbf{EmbSim} &43.6  &51.6 &33.0 &34.0 &65.0 &64.7 &65.6 &72.0 &49.1 &42.3\\
    \rowcolor{red!10}\textbf{GraphSim} &\textbf{47.8}  &\textbf{53.4} &\textbf{35.0} &\textbf{35.0}&\textbf{68.5} &\textbf{73.3} &\textbf{68.8} &\textbf{74.4} &\textbf{51.3} &\textbf{42.5}\\
    \rowcolor{cyan!10}\textbf{$\mathcal{L}_{LLM}$} &45.6  &50.4 &31.0 &34.2 &52.8 &63.7 &64.0  &74.2 &\underline{41.5} &50.1\\
    \rowcolor{cyan!10}\textbf{GLIP} &\textbf{46.6}  &\textbf{50.8} &\textbf{33.0} &\textbf{35.2} &\textbf{56.3} &\textbf{68.9}  &\textbf{66.6} &\textbf{75.0} &\textbf{41.5}  &\textbf{51.7}\\
    \rowcolor{cyan!10}\textbf{Ours} &\underline{46.2}  &\underline{50.4} &\underline{32.6} &\underline{34.6} &\underline{54.8} &\underline{66.9} &\underline{65.6}&\underline{74.5} &41.3 &\underline{51.3}\\ \hdashline
    \rowcolor{cyan!10}\textbf{Oracle} &48.0  &51.2 &34.0 &36.2 &56.5 &70.9 &66.6 &75.2 &41.7 &52.7\\
    \rowcolor{gray!20} \multicolumn{11}{c}{\textbf{\emph{LLaMA2-13B}}}\\ 
    \textbf{Zero-shot} &52.0  &52.0 &9.2 &9.2 &65.4 &65.4  &55.8 &55.8 &55.3 &55.3\\ 
    \rowcolor{red!10}\textbf{EmbSim} &59.2  &66.0 &39.0 &21.6 &77.2 &76.6 &83.4 &84.6 &63.7 &49.1\\
    \rowcolor{red!10}\textbf{GraphSim} &\textbf{62.6}  &\textbf{66.4} &\textbf{39.4} &\textbf{25.6} &\textbf{79.4} &\textbf{83.0} &\textbf{84.8} &\textbf{84.8} &\textbf{64.5} &\textbf{53.9}\\
    \rowcolor{cyan!10}\textbf{$\mathcal{L}_{LLM}$} &62.6  &66.2 &38.0 &38.6 &55.3 &77.6 &82.6 &85.8 &62.3 &63.7\\
    \rowcolor{cyan!10}\textbf{GLIP} &\textbf{63.8}  &\textbf{67.2} &\textbf{39.8} &\textbf{41.4} &\textbf{72.4} &\textbf{86.6} &\underline{83.2} &\underline{86.2} &\textbf{62.7} &\underline{63.7}\\
    \rowcolor{cyan!10}\textbf{Ours} &\underline{63.4}  &\underline{67.2}  &\underline{38.6} &\underline{41.0} &\underline{71.4} &\underline{86.2} &\textbf{83.2} &\textbf{86.2} &\underline{62.5} &\textbf{63.7}\\ \hdashline
    \rowcolor{cyan!10}\textbf{Oracle} &64.6  &67.6 &41.0 &43.0 &71.6 &87.2 &84.4 &87.0 &62.7 &63.8\\
    \rowcolor{gray!20} \multicolumn{11}{c}{\textbf{\emph{GPT-3.5}}}\\ 
    \textbf{Zero-shot} &74.6  &74.6 &49.6 &49.6 &57.5 &57.5 &91.2 &91.2 &74.0 &74.0\\ 
    \rowcolor{red!10}\textbf{EmbSim} &78.2  &\textbf{81.2} &50.0 &53.2 &81.2 &93.6 &92.2 &\textbf{94.2} &74.2 &72.4\\
    \rowcolor{red!10}\textbf{GraphSim} &\textbf{81.6}  &79.6 &\textbf{55.2} &\textbf{53.6} &\textbf{93.6} &\textbf{94.4} &\textbf{95.6} &94.0 &\textbf{74.4} &\textbf{72.6}\\ 
    \rowcolor{cyan!10}\textbf{$\mathcal{L}_{LLM}$} &82.0  &80.8 &58.8 &60.8 &76.4 &76.0 &\underline{94.4} &94.4 &72.0  &\textbf{76.0}\\
    \rowcolor{cyan!10}\textbf{GLIP} &\textbf{83.2} &\textbf{84.0} &\textbf{60.0} &\textbf{63.2} &\textbf{83.2} &\textbf{78.4} &\underline{95.2} &\textbf{94.4} &\textbf{75.2} &\underline{74.8} \\
    \rowcolor{cyan!10}\textbf{Ours} &\underline{82.6}  &\underline{82.6} &\underline{59.2} &\underline{62.0} &\underline{81.2} &\underline{77.2} &\textbf{95.4} &94.2 &\underline{73.4} &74.6\\ \hdashline
    \rowcolor{cyan!10}\textbf{Oracle} &82.4 &83.2 &60.4 &61.6 &85.6 &91.2 &96.8 &96.4 &73.6 &77.2\\

\bottomrule[1.2pt]
\end{tabular}
\label{tab:main-results}
\end{table*}

\subsection{Ablation Study of GraphSim Components}

\begin{figure}[t]
\centering
\includegraphics[width=\linewidth]{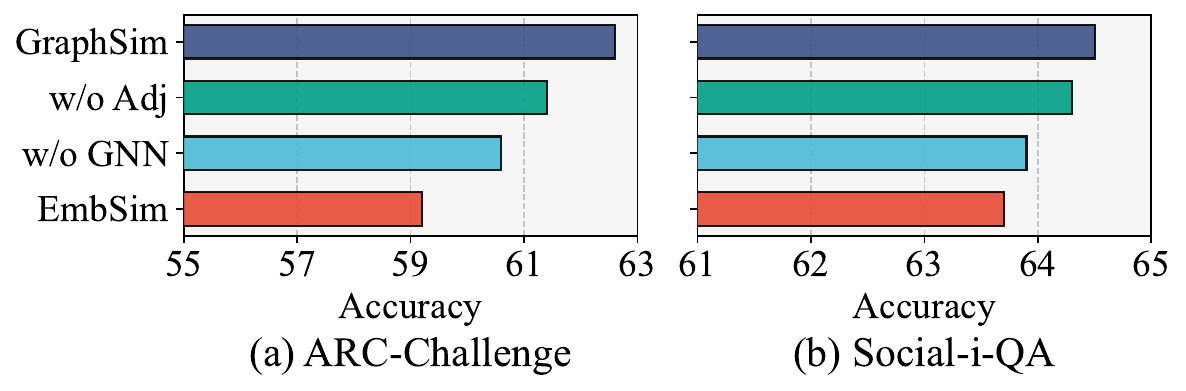}
\caption{Ablation Study of GraphSim Components.}
\label{fig:graphsim}
\vspace{-0.2in}
\end{figure}

We conduct an ablation study to assess the individual contributions of GraphSim’s key components to overall performance. Specifically, we evaluate GraphSim on two tasks: ARC-Challenge and Social-i-QA, using LLaMA2-13B under the one-shot in-context learning setting. As shown in Figure~\ref{fig:graphsim}, we examine the effects of removing (i) adjacency-based aggregation (GraphSim w/o Adj) and (ii) GNN-based aggregation (GraphSim w/o GNN). Note that removing both components reduces GraphSim to EmbSim.

The results show that excluding GNN-based aggregation leads to a significant performance drop across both tasks, underscoring the importance of capturing multiple structural perspectives for robust cross-task example selection. Overall, both components contribute meaningfully to the effectiveness of GraphSim. In Section~\ref{sec:moregnn}, we provide an empirical analysis of the influence of the number of GNNs in GNN-based aggregation

\subsection{Influence of Number of ICL Examples}

\begin{figure}[htbp]
\centering
\includegraphics[width=\linewidth]{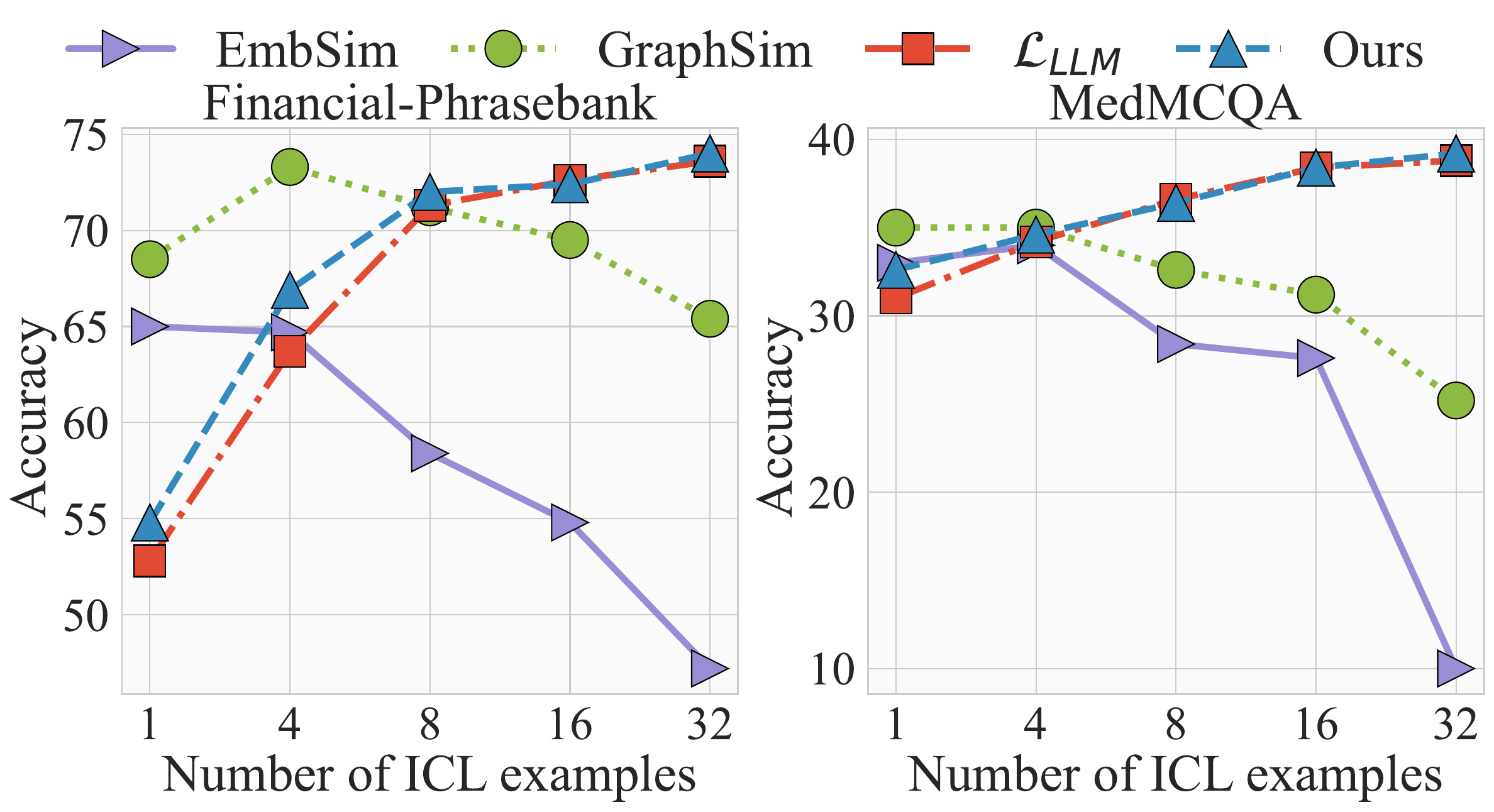}
\caption{Accuracy variation with respect to the number of ICL examples using LLaMA2-7B.}
\label{fig:diffk}
\vspace{-0.2in}
\end{figure}

In this subsection, we investigate how the number of in-context examples affects ICL performance. Specifically, we compare our pipeline with two cross-task ICL baselines: EmbSim and GraphSim, and one in-task ICL baseline $\mathcal{L}_{\text{LLM}}$. Specifically, we evaluate methods on two tasks: Financial-Phrasebank and MedMCQA, using LLaMA2-7B under the different shot ICL setting. The results are presented in Figure~\ref{fig:diffk}.

We observe that while cross-task ICL methods may perform well in low-shot settings (e.g., one-shot or four-shot), their performance degrades as the number of examples increases. This decline is likely due to distribution shifts and label space mismatches between source and target tasks. Adding more cross-task examples introduces noise that can confuse the LLM. This observation is consistent with the findings of \citet{chatterjee2024language}. Nevertheless, GraphSim demonstrates greater robustness than EmbSim, due to its incorporation of structural information that generalizes better across tasks.

On the other hand, in-task ICL methods benefit from an increasing number of examples. Our pipeline achieves performance comparable to $\mathcal{L}_{\text{LLM}}$, while being more cost-efficient, as it relies less on expensive LLM calls.

\subsection{Discussion of Labeling Efficiency}
In this section, we analyze the labeling efficiency of different methods. Specifically, we measure the total runtime for pseudo-labeling and inference on 500 test samples from the MedMCQA task using the LLaMA2-7B model on an NVIDIA A6000 GPU. Results are presented in Table~\ref{tab:efficiency}.

We observe that the runtime of $\mathcal{L}_{\text{LLM}}$ increases approximately linearly with the size of $\gD^U$, as each additional sample requires an LLM call. This results in a significant increase in overall cost, especially when using commercial API-based LLMs,and the issue is further exacerbated with larger models due to their higher inference latency.

While our full pipeline introduces some overhead compared to GLIP alone (due to the additional GraphSim step for pseudo-labeling $\gD^L$), it only requires LLM calls on a small subset of samples. As a result, the total cost grows much more slowly with the size of $\gD^U$, making our method a more efficient and scalable solution for pseudo-labeling in practice.

\begin{table}[t]
\caption{Total runtime (in minutes) for pseudo-labeling and inference on 500 MedMCQA test samples using LLaMA2-7B. We vary the sizes of $\gD^L$ and $\gD^U$; for example, "100/400" means $|\gD^L|=100$, $|\gD^U|=400$.}
\centering
\small
\tabcolsep = 4pt
\begin{tabular}{l c c c c}
\toprule[1.2pt]
\textbf{Method} & \textbf{100/400} & \textbf{200/400} & \textbf{400/800} & \textbf{100/1600}\\ 
\midrule
\textbf{GLIP} &3.0  &3.0   &3.2  &3.5    \\ 
\textbf{Ours} &3.4  &3.8   &4.3  &4.9    \\ 
$\mathcal{L}_{\text{LLM}}$ &3.6  &3.6   &5.3  &8.4   \\ 
\bottomrule[1.2pt]
\end{tabular}
\vspace{-0.2in}
\label{tab:efficiency}
\end{table}

\section{Ablation Study of Different Number of GNNs in GraphSim}\label{sec:moregnn}

We conduct ablation experiments to analyze how the number of GNNs in GraphSim affects performance. The evaluation is carried out on three target tasks under one-shot settings using the LLaMA2-13B model. Results are shown in Table~\ref{tab:gnn}.

We observe that using four GNNs, two with a single layer and two with two layers, yields the best average performance. Using fewer GNNs reduces the model’s ability to capture structural information from diverse perspectives. On the other hand, increasing the number of GNNs leads to performance degradation due to the curse of dimensionality, as the final embedding size grows linearly with the number of GNNs, making similarity computation less meaningful in high-dimensional space.
These results suggest that selecting an appropriate number of GNNs is crucial for balancing expressiveness and representational efficiency in cross-task similarity computation.

\begin{table}[t]
\caption{Performance of GraphSim under the one-shot setting using LLaMA2-13B. We vary the number and depth of GNNs; for example, "[1,1,2,2]" indicates two GNNs with one layer and two with two layers.}
\centering
\small
\tabcolsep = 4pt
\begin{tabular}{l c c c}
\toprule[1.2pt]
\textbf{GNN Setting} & \textbf{ARC-Challenge} & \textbf{MedMCQA} & \textbf{SciQ}\\ 
\midrule
\textbf{[1,2]} &61.2  &\textbf{39.6}   &\underline{84.6}    \\ 
\textbf{[1,1,2,2]} &\textbf{62.6}  &\underline{39.4}   &\textbf{84.8}      \\ 
\textbf{[1,1,1,2,2,2]} &\underline{61.6}  &38.6    &84.2   \\ 
\bottomrule[1.2pt]
\end{tabular}
\label{tab:gnn}
\end{table}

\section{Conclusions}
We propose a cost-effective and scalable pipeline for pseudo-labeling novel target tasks via cross-task ICL. we introduce GraphSim, a graph-based method that incorporates structural information to improve cross-task similarity estimation. To further reduce LLM reliance, we develop GLIP, a graph-based label propagation technique that extends a small pseudo-labeled seed set to the full target dataset without additional LLM queries.
Our approach combines the generalizability of cross-task supervision with the efficiency of structure-aware label propagation. Experiments across multiple NLP benchmarks show that our method achieves competitive ICL performance while significantly lowering annotation and inference costs.
We hope this work encourages future research on graph-enhanced ICL and promotes practical solutions for applying LLMs in low-resource scenarios.

\section{Limitations}

While our work provides an efficient pipeline for cross-task pseudo-labeling to facilitate in-context learning on novel tasks, there remain several promising avenues for future research.

\begin{itemize}[left=0pt]
    \item \textit{Focus on Cross-Task Pseudo-Labeling}: Our goal is to construct demonstration pools for unseen target tasks by leveraging labeled examples from well-established source tasks. The current framework assumes access to one suitable source task, but identifying the best source task for a given target remains an open question. A promising direction is to measure the similarity between source and target task definitions, for instance, using the final-layer hidden representations of their prompts~\cite{chatterjee2024language}. We leave a systematic study of source task selection strategies to future work.

   \item \textit{Applicability Beyond Multiple-Choice QA}: We primarily focus on multiple-choice tasks in this work. Although certain generative or reasoning tasks can be reformulated into multiple-choice formats—e.g., by prompting an LLM to generate plausible distractors—such transformations introduce additional cost and complexity. Investigating the effectiveness and efficiency of our pipeline in these reformulated settings is left for future work.

   \item \textit{Mixed-Task In-Context Learning}: Prior works such as \citet{chatterjee2024language} explore mixed-task ICL, where demonstrations are drawn from multiple tasks. Our study focuses instead on the single-source-task setting to investigate how to best exploit one available source for cross-task labeling. Extending our method to handle multi-source or mixed-task scenarios is an important direction for future research.
\end{itemize}

\section{Ethics Statement}

Our work focuses on developing a cost-efficient cross-task ICL labeling pipeline using publicly available datasets and pretrained language models. While acknowledging the need for responsible usage of the proposed method, we do not foresee major negative societal impacts.

\section*{Acknowledgments}
This work is supported in part by the National Science Foundation (NSF) under grants IIS-2006844, IIS-2144209, IIS-2223769, CNS-2154962, BCS-2228534, CMMI-2411248, ECCS-2143559, and CPS-2313110; the Office of Naval Research (ONR) under grant N000142412636; and the Commonwealth Cyber Initiative (CCI) under grant VV-1Q24-011.

\bibliography{custom}

\clearpage
\appendix

\section{Dataset details}
\label{sec:appendix-dataset}
In this section, we provide detailed descriptions of the source and target datasets used in our experiments, following the setup of \citet{chatterjee2024language}. We also include the task-specific instructions used in prompting for each dataset. 


\subsection{Source datasets}
We have used the following datasets as source datasets: 
\begin{itemize}
    \item[] \textbf{ARC-Easy:} ARC-Easy \cite{Clark2018ThinkYH} is a subset of the ARC (AI2 Reasoning Challenge) dataset containing multiple-choice science questions for 3rd–9th grade students. Each question has four options, with one correct answer. We use the 2,251-question training set as the source dataset.
    
    \item[] \textbf{AG-news:} AG-news \cite{Zhang2015CharacterlevelCN} is a news classification dataset with articles categorized into four classes: sports, business, technology, and world. We randomly sample 10K articles from the 120K training set for source examples.
    
    \item[] \textbf{BoolQ:} BoolQ \cite{clark-etal-2019-boolq} is a reading comprehension dataset with yes/no questions based on associated passages. The 9,427 labeled question-passage pairs are used as source examples.
    
    \item[] \textbf{Commonsense-QA:} Commonsense-QA \cite{talmor-etal-2019-commonsenseqa} is a multiple-choice QA dataset requiring commonsense reasoning. Each question has five options, one of which is correct. We sample source examples from the 9,740-question training set.

    \item[] \textbf{QQP:} Quora Question Pairs (QQP) \cite{DBLP:journals/corr/abs-1907-01041} dataset is curated for the task of natural language understanding. This dataset consists of question pairs collected from the popular question-answering website {\em Quora}, and the task is to determine if the questions are duplicates of each other. We sample 10K labeled pairs for source usage.

    \item[] \textbf{RACE:} RACE \cite{lai-etal-2017-race} is a reading comprehension dataset sourced from English exams for students aged 12–18, with multiple-choice questions based on passages. We sample 10K passage-question pairs from the 87.9K available in the training set.
    
    \item[] \textbf{SST2:} SST2 \cite{socher-etal-2013-recursive} is a sentiment classification dataset from the Stanford Sentiment Treebank, where each movie review is labeled as positive or negative. We use 10K samples from the 67.3K training examples.
    
\end{itemize}
\begin{table*}[t]
\caption{Additional experiment results using two recent LLMs, LLaMA3-8B and GPT-4o. Dashed lines separate cross-task ICL results from in-task ICL results. Bold font highlights the best performance within each category, while underlined values indicate the second-best results within the in-task ICL group for clearer analysis.}
\centering
\small
\tabcolsep = 4pt
\begin{tabular}{l c c c c c c c c c c}
\toprule[1.2pt]
    \multirow{2}{*}{\textbf{Method}} & \multicolumn{2}{c}{\textbf{ARC-Challenge}} & \multicolumn{2}{c}{\textbf{MedMCQA}}& \multicolumn{2}{c}{\textbf{Financial-Phrasebank}}& \multicolumn{2}{c}{\textbf{SciQ}}& \multicolumn{2}{c}{\textbf{Social-i-QA}} \\ 
    \cmidrule(lr){2-3} \cmidrule(lr){4-5} \cmidrule(lr){6-7} \cmidrule(lr){8-9} \cmidrule(lr){10-11}
     &$K=1$ &$K=4$	&$K=1$ &$K=4$ &$K=1$ &$K=4$ &$K=1$ &$K=4$ &$K=1$ &$K=4$  \\ 
    \midrule
    \rowcolor{gray!20} \multicolumn{11}{c}{\textbf{\emph{LLaMA3-8B}}}\\ 
    \textbf{Zero-shot} &6.4  &6.4 &3.2 &3.2 &73.2 &73.2 &2.4 &2.4 &4.8 &4.8\\ 
    \rowcolor{red!10}\textbf{EmbSim} &80.0  &79.2 &57.6 &56.2 &69.6 &71.8 &90.4 &90.8 &72.8 &65.2\\
    \rowcolor{red!10}\textbf{GraphSim} &\textbf{81.0}  &\textbf{80.0} &\textbf{60.0} &\textbf{57.2} &\textbf{72.4} &\textbf{73.2} &\textbf{90.6} &\textbf{92.8} &\textbf{73.2} &\textbf{72.4}\\
    \rowcolor{cyan!10}\textbf{$\mathcal{L}_{LLM}$} &78.4  &80.4 &\underline{57.2} &60.0 &\underline{82.2} &\textbf{86.8} &89.6 &91.6 &\underline{71.2} &\textbf{72.4}\\
    \rowcolor{cyan!10}\textbf{GLIP} &\underline{79.6}  &\textbf{82.4} &\textbf{57.2} &\underline{60.0} &\textbf{82.4} &86.4 &\textbf{92.4} &\underline{92.4} &\textbf{71.4} &71.2\\
    \rowcolor{cyan!10}\textbf{Ours} &\textbf{80.0}  &\underline{81.2} &56.6 &\textbf{60.0} &82.0 &\underline{86.6} &\underline{91.8} &\textbf{92.4} &70.6 &\underline{71.4}\\ \hdashline
    \rowcolor{cyan!10}\textbf{Oracle} &81.6  &82.0 &57.6 &59.6 &86.0 &92.4 &92.8 &93.6 &73.2 &71.6\\
    \rowcolor{gray!20} \multicolumn{11}{c}{\textbf{\emph{GPT-4o}}}\\ 
    \textbf{Zero-shot} &93.6  &93.6 &73.2 &73.2 &71.6 &71.6 &98.4 &98.4 &82.8 &82.8\\ 
    \rowcolor{red!10}\textbf{EmbSim} &92.0  &93.2 &73.8 &72.6 &86.3 &\textbf{89.2} &98.2 &97.6 &\textbf{81.2} &80.3\\
    \rowcolor{red!10}\textbf{GraphSim} &\textbf{94.2}  &\textbf{94.4} &\textbf{74.4} &\textbf{73.5} &\textbf{88.6} &88.9 &\textbf{98.4} &\textbf{97.9} &80.9 &\textbf{82.1}\\ 
    \rowcolor{cyan!10}\textbf{$\mathcal{L}_{LLM}$} &96.0  &\underline{96.0} &\textbf{75.2} &\underline{78.4} &\underline{93.6} &\textbf{92.8} &96.6 &\underline{98.8} &\textbf{81.6} &82.4\\
    \rowcolor{cyan!10}\textbf{GLIP} &\textbf{97.2}  &\textbf{96.6} &74.8 &\textbf{78.8} &\textbf{94.0} &\underline{92.2} &\textbf{98.4} &\textbf{98.8} &\underline{80.8} &\textbf{83.6} \\
    \rowcolor{cyan!10}\textbf{Ours} &\underline{96.2}  &95.2 &\underline{75.0} &78.0 &92.8 &91.8 &\underline{97.5} &97.9 &80.3 &\underline{83.0}\\ \hdashline
    \rowcolor{cyan!10}\textbf{Oracle} &95.6 &96.4   &77.6 &78.4  &96.8 &96.4 &98.0 &98.8 &83.4 &84.2\\

\bottomrule[1.2pt]
\end{tabular}
\label{tab:additional-llm-results}
\end{table*}
\subsection{Target datasets}
We have used the following datasets as target datasets:

\begin{itemize}
    \item[] \textbf{ARC-Challenge:} ARC-Challenge \cite{Clark2018ThinkYH} is a subset of the ARC (AI2 Reasoning Challenge) dataset containing difficult science questions for students in grades 3–9. These are questions that were incorrectly answered by both a retrieval-based system and a word co-occurrence method, making them more challenging. Each question is multiple-choice with four options, one of which is correct. We randomly select 500 questions from the 1,172-question test set for our target dataset.

    \item[] \textbf{Social-i-QA:} Social-i-QA \cite{DBLP:journals/corr/abs-1904-09728} is a commonsense reasoning dataset focused on social and emotional understanding. Each example presents a social scenario and a multiple-choice question with three options. We sample 500 examples from the 1,954 available in the validation set for use as our target data.

    \item[] \textbf{SciQ:} SciQ \cite{SciQA2023} is a multiple-choice science QA dataset covering topics in physics, chemistry, and biology. Each question has four answer choices. We construct our target dataset by sampling 500 examples from the 1,000-question test set.

    \item[] \textbf{MedMCQA:} MedMCQA \cite{pmlr-v174-pal22a} is a multiple-choice QA dataset derived from Indian postgraduate medical entrance exam questions. Each question has four options, with one correct answer. We sample 500 questions from the 4,183-question validation set to form our target dataset.
    
    \item[] \textbf{Financial-Phrasebank:} Financial-Phrasebank \cite{Malo2014GoodDO} is a sentiment analysis dataset consisting of 4,840 sentences from English-language financial news, categorized by sentiment. The annotators were instructed to assess the sentences from an investor's perspective, determining whether the news would likely have a positive, negative, or neutral impact on stock prices.

\end{itemize}

In each case, the selection, though random, is done in such a way that our target datasets are balanced, i.e. the number of examples with each of the different possible labels is almost equal.


\section{Additional Experiments on More LLMs}\label{sec:morellm}

In this section, we evaluate our pipeline on two additional large language models: LLaMA3-8B~\cite{llama3modelcard} and GPT-4o~\cite{hurst2024gpt}. Note that these models were not included in the experiments of \citet{chatterjee2024language}. For cross-task ICL, we pair LLaMA3-8B with the same source-target settings used for LLaMA2-7B, and use the same GPT-3.5 settings for GPT-4o. While this may not yield the optimal source task selection for each model, it still allows us to analyze overall performance trends.

The results, shown in Table~\ref{tab:additional-llm-results}, indicate that GraphSim consistently outperforms EmbSim across different tasks, reaffirming its effectiveness for cross-task example selection. In the in-task setting, we observe that $\mathcal{L}_{LLM}$ achieves performance comparable to our pipeline when using GPT-4o. This is largely due to the strong capabilities of GPT-4o; however, $\mathcal{L}_{LLM}$ also incurs higher labeling costs. Overall, our pipeline continues to provide a cost-efficient solution for generating high-quality pseudo-labels on novel target tasks, even when compared to stronger LLMs.

\section{Additional Comparison with Demonstration-generation ICL Methods.}\label{sec:morebaseline}
\begin{table}[t]
\caption{Comparison with demonstration-generation ICL baselines.}
\centering
\small
\tabcolsep = 2pt
\begin{tabular}{l c c c}
\toprule[1.2pt]
\textbf{Method} & \textbf{MedMCQA} & \textbf{Financial-Phrasebank}  & \textbf{Social-i-QA}\\ 
\midrule
\textbf{Self-ICL} &11.2  &5.2   &27.6   \\ 
\textbf{SG-ICL} &15.4  &6.2   &28.0      \\ 
\textbf{Ours} &34.6  &66.9 &51.3   \\ 
\bottomrule[1.2pt]
\end{tabular}
\label{tab:selficl}
\end{table}
In addition to cross-task examples and zero-shot prompting, recent research has explored prompting LLMs to generate pseudo-demonstrations for a given task, which are then used for ICL at test time. In this section, we compare our proposed approach with two domain-agnostic demonstration-generation methods: Self-ICL~\cite{chen2023self} and SG-ICL~\cite{kim2022self}. We use LLaMA2-7B as the base LLM in a 4-shot setting, with results shown in Table~\ref{tab:selficl}.

Both baseline methods show limited performance on the target tasks, primarily because the LLM struggles to understand the task from the description alone, consistent with the weak performance observed in the zero-shot ICL setting. As a result, these methods are unable to generate high-quality demonstrations, highlighting the importance of better example selection strategies like ours.

\section{Ablation study.}\label{sec:ablation}

In this section, we present an ablation study on the hyperparameter $\lambda$, which balances the training objective for the GNN. We experiment with $\lambda$ values ranging from 0.2 to 1.0 on the Financial-Phrasebank and SciQ tasks. We report 4-shot performance using GLIP with LLaMA2-13B as the base LLM. The results are shown in Table~\ref{tab:lambda}.

Performance remains relatively stable across different values of $\lambda$, demonstrating the robustness of our method to this hyperparameter. While tuning $\lambda$ on a development set could yield marginal improvements, it would also require substantially more LLM queries. To balance performance and computational cost, we adopt a fixed value that already delivers strong results.

\begin{table}[t]
\caption{Influence of $\lambda$.}
\centering
\small
\tabcolsep = 4pt
\begin{tabular}{l c c c c c}
\toprule[1.2pt]
\textbf{Dataset} & \textbf{0.2} & \textbf{0.4}  & \textbf{0.6} & \textbf{0.8} & \textbf{1.0}\\ 
\midrule
\textbf{Financial-Phrasebank} &86.4	&86.6	&\textbf{86.8}	&85.6	&85.2\\
\textbf{SciQ} &85.4	&\textbf{86.2}	&86.0	&85.8	&85.0\\
\bottomrule
\end{tabular}
\label{tab:lambda}
\end{table}

\begin{table*}[!t]
\small
\begin{center}
\scalebox{1}{
\begin{tabular}{|p{0.2\textwidth}|p{0.75\textwidth}|}
 \hline
 {\bf Source task}&{\bf Task definition} \\
  \hline

AG-news & Given a sentence do text classification, the sentence is a clipping from a news article that may be either related to sports, business, technology, or world news. You are to recognize the category of the sentence and label them as "sports", "business", "technology" or "world" news 
 \\
\hline

ARC-Easy & Given a question answering task from the 3rd to 9th-grade science exam. The question contains four options "A.", "B.", "C." and "D." Select the most appropriate choice that answers the question
\\
\hline

BoolQ & Given a context and a question do binary true and false type text classification. You are given a passage as context and a question related to the passage that can be answered as "True" or "False". Based on the context, question and your reasoning ability answer in a "True" and "False".
\\
\hline

Commonsense-QA &The following task relates to commonsense reasoning. It consists of a question that can be easily solved using logical abilities and reasoning, a set of five options  "A.", "B.", "C.", "D." and "E." are also provided along with the question, one of these options answers the question logically. Use your reasoning ability to select the most appropriate answer from the provided choices "A.", "B.", "C.", "D." and "E." and assign these choices (i.e  "A.", "B.", "C.", "D." and "E.") as the label 

\\
\hline




QQP &Given two question pairs do text classification based on whether they are duplicates or not. The questions are mined from the popular online discussion forum Quora. As duplicate quetion might be present on Quora, the task is to label two identical questions as "duplicate" if they ask the same query else label the pair as "not duplicate".
\\
\hline

RACE &Given a reading comprehension type question-answering from an english exam for school students. You are given a context and multiple choice question containing four options "A.", "B.", "C." and "D.". The question is answerable from the comprehension. Based on the question, the option and the context select the most appropriate answer from the provided choices "A.", "B.", "C." and "D.".
\\
\hline

SST2 &Given a movie review do text classification, based on the sentiment conveyed by the review label it as "positive" or "negative"
\\
\hline

\end{tabular}
}
\caption{Task definitions of source tasks}
\vspace{-2mm}
\label{tab:source-task-definitions}
\end{center}
\end{table*}

\begin{table*}[!t]
\small
\begin{center}
\scalebox{1}{
\begin{tabular}{|p{0.2\textwidth}|p{0.75\textwidth}|}
 \hline
 {\bf Target task}&{\bf Task definition} \\
  \hline

ARC-Challenge & Given a question answering task from the 3rd to 9th-grade science exam. The question contains four options "A.", "B.", "C." and "D." Select the most appropriate choice that answers the question\\
\hline

Financial-Phrasebank & Given a sentence mined from a financial news article, you are to determine the sentiment polarity of the sentence. The task deals with financial sentiment analysis. Based on the sentiment conveyed by the sentence, label the sentence as "negative", "positive" or "neutral"\\
\hline

MedMCQA & Given a multiple choice question containing four options "A.", "B.", "C." and "D." from a medical entrance exam. The question is related to a sub-field of medical science like Microbiology, Radiology, Ophthalmology, Surgery, Human anatomy, etc. Based on the question, the option and your knowledge of the medical field select the most appropriate answer from the provided choices "A.", "B.", "C." and "D.".\\
\hline

SciQ & Given a question from a scientific exam about Physics, Chemistry, and Biology, among others. The question is in multiple choice format with four answer options "A.", "B.", "C." and "D.". Using your knowledge about the scientific fields answer the question and provide the label "A", "B", "C" and "D" as answer
\\
\hline

Social-i-QA & Given an action as the context and a related question, you are to answer the question based on the context using your social intelligence. The question is of multiple choice form with three options "A", "B" and "C". Select the most appropriate answer from the provided choices "A", "B" and "C".
\\
\hline
\end{tabular}
}
\caption{Task definitions of target tasks}
\vspace{-2mm}
\label{tab:target-task-definitions}
\end{center}
\end{table*}
\end{document}